\documentclass[review]{elsarticle}
\usepackage{geometry}[a4paper, scale=0.67]

\usepackage{amssymb}
\usepackage{lineno,hyperref}
\usepackage{bm}
\usepackage[flushleft]{threeparttable}
\usepackage{multirow}

\usepackage{tabularx}
\usepackage{float}
\usepackage{subfig}

\usepackage{makecell}
\usepackage{pifont}
\usepackage{booktabs}

\usepackage{amsfonts}
\usepackage{amsmath}

\journal{arXiv}

\begin{document}

\begin{frontmatter}



\title{Do We Fully Understand Students' Knowledge States? Identifying and Mitigating Answer Bias in Knowledge Tracing}


\author[1]{Chaoran~Cui}
\ead{crcui@sdufe.edu.cn}
\author[1]{Hebo~Ma}
\ead{heboma98@gmail.com}
\author[2]{Chen~Zhang}
\ead{jason-c.zhang@polyu.edu.hk}
\author[1]{Chunyun~Zhang}
\ead{zhangchunyun1009@126.com}
\author[1]{Yumo~Yao}
\ead{ymyao@mail.sdufe.edu.cn}
\author[3]{Meng~Chen}
\ead{mchen@sdu.edu.cn}
\author[4]{Yuling~Ma}
\ead{mayuling20@sdjzu.edu.cn}

\address[1]{School of Computer Science and Technology, Shandong University of Finance and Economics, Jinan, China}
\address[2]{Department of Computing, Hong Kong Polytechnic University, Hong Kong, China}
\address[3]{School of Software, Shandong University, Jinan, China}
\address[4]{School of Computer Science and Technology, Shandong Jianzhu University, Jinan, China}

% \cortext[cor1]{Corresponding author: Chaoran Cui}

\begin{abstract}
Knowledge tracing (KT) aims to monitor students' evolving knowledge states through their learning interactions with concept-related questions, and can be indirectly evaluated by predicting how students will perform on future questions.
In this paper, we observe that there is a common phenomenon of answer bias, i.e., a highly unbalanced distribution of correct and incorrect answers for each question.
Existing models tend to memorize the answer bias as a shortcut for achieving high prediction performance in KT, thereby failing to fully understand students' knowledge states.
To address this issue, we approach the KT task from a causality perspective.
A causal graph of KT is first established, from which we identify that the impact of answer bias lies in the direct causal effect of questions on students' responses.
A novel COunterfactual REasoning (CORE) framework for KT is further proposed, which separately captures the total causal effect and direct causal effect during training, and mitigates answer bias by subtracting the latter from the former in testing.
The CORE framework is applicable to various existing KT models, and we implement it based on the prevailing DKT, DKVMN, and AKT models, respectively.
Extensive experiments on three benchmark datasets demonstrate the effectiveness of CORE in making the debiased inference for KT.
We have released our code at \url{https://github.com/lucky7-code/CORE}.
\end{abstract}



\end{frontmatter}

% \linenumbers

\section{Introduction}
The COVID-19 pandemic outbreak has compelled millions of students to stay home and pursue their studies online in recent years.
%In online learning environments, it is important to automatically estimate students' \emph{knowledge states}, i.e., their mastery levels of different knowledge concepts (e.g., the concepts of addition, subtraction, and multiplication in math)~\cite{Abdelrahman2022Knowledge}.
In online learning environments, it is crucial to automatically assess students' \emph{knowledge states}, which refers to their mastery levels of various knowledge concepts, such as addition, subtraction, and multiplication in math~\cite{abdelrahman2023learning}.
% This feature could enable personalized learning experiences to be provided for students, such as learning resource recommendation~\cite{wang2022personalized}, adaptive testing~\cite{ghosh2021BOBCAT}, and educational gaming~\cite{long2017educational}.
By leveraging this functionality, it becomes possible to offer tailored learning experiences to students, including learning resource recommendation~\cite{lin2022hierarchical}, adaptive testing~\cite{ghosh2021BOBCAT}, and academic performance prediction~\cite{cui2022tri}.

% levels of students mastering different knowledge concepts (e.g., the concepts of linear equation, power function, and inequality in mathematics), which are termed by \emph{knowledge states}.

\begin{figure}
    \centering
    \includegraphics[width=0.7\textwidth]{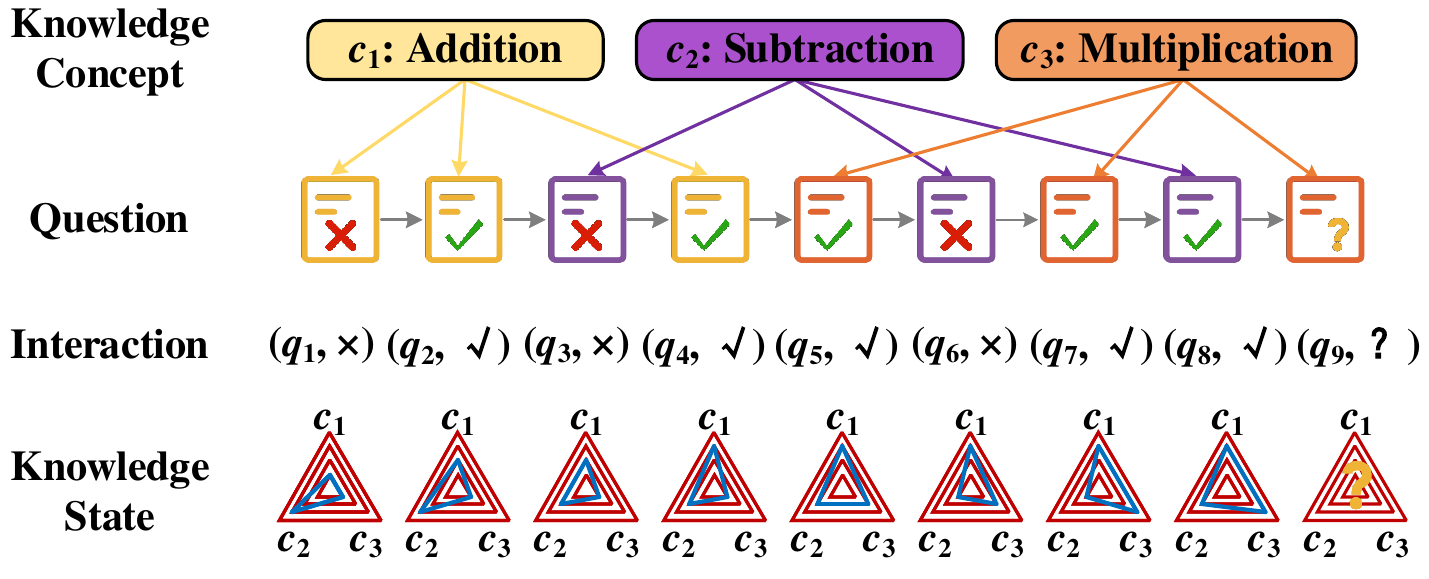}
    \caption{A simple schematic diagram of KT. Given a learning sequence of students on concept-related questions, KT aims to monitor the students’ knowledge states over time. This can be indirectly evaluated by predicting the students’ performance on future questions.
    Note that KT models consider the students' performance to be a binary outcome, typically based on the correctness of their answers to questions.}
    \label{fig:KT}
\end{figure}

% Knowledge tracing (KT) aims to monitor students' evolving knowledge states based on the sequence of their responses to concept-related exercises~\cite{Abdelrahman2022Knowledge,liu2021survey}.
% Knowledge tracing (KT) aims to track students' evolving knowledge states through their learning interactions with concept-related exercises~\cite{Abdelrahman2022Knowledge,liu2021survey}.
Knowledge tracing (KT) aims to track students' evolving knowledge states through their engagement with concept-related questions~\cite{Abdelrahman2022Knowledge,song2022survey}.
As shown in Fig.~\ref{fig:KT}, students are presented with a sequence of questions related to a set of knowledge concepts and are required to provide answers to these questions.
Throughout the learning process, a KT system dynamically assesses the students' knowledge states over the concepts.
%, so that we can predict how students will perform on future interactions.
% Generally, KT simplifies the interaction process by assuming students' responses are only binary, i.e., correct or incorrect answers.
In the literature, KT is commonly framed as a sequence modeling problem and has been extensively investigated for decades~\cite{corbett1994knowledge,khajah2016deep}.
With the advent of deep learning, deep knowledge tracing (DKT)~\cite{piech2015deep} was proposed to apply recurrent neural networks (RNNs) to capture the long-term dependencies among a student's learning interactions.
After that, more types of neural networks were developed for KT, such as the dynamic key-value memory networks (DKVMN)~\cite{zhang2017dynamic} and self-attentive knowledge tracing networks (AKT)~\cite{ghosh2020context}.

\begin{figure}[t]
    \centering
    \subfloat[ASSIST09]
    {\includegraphics[width=0.325\textwidth]{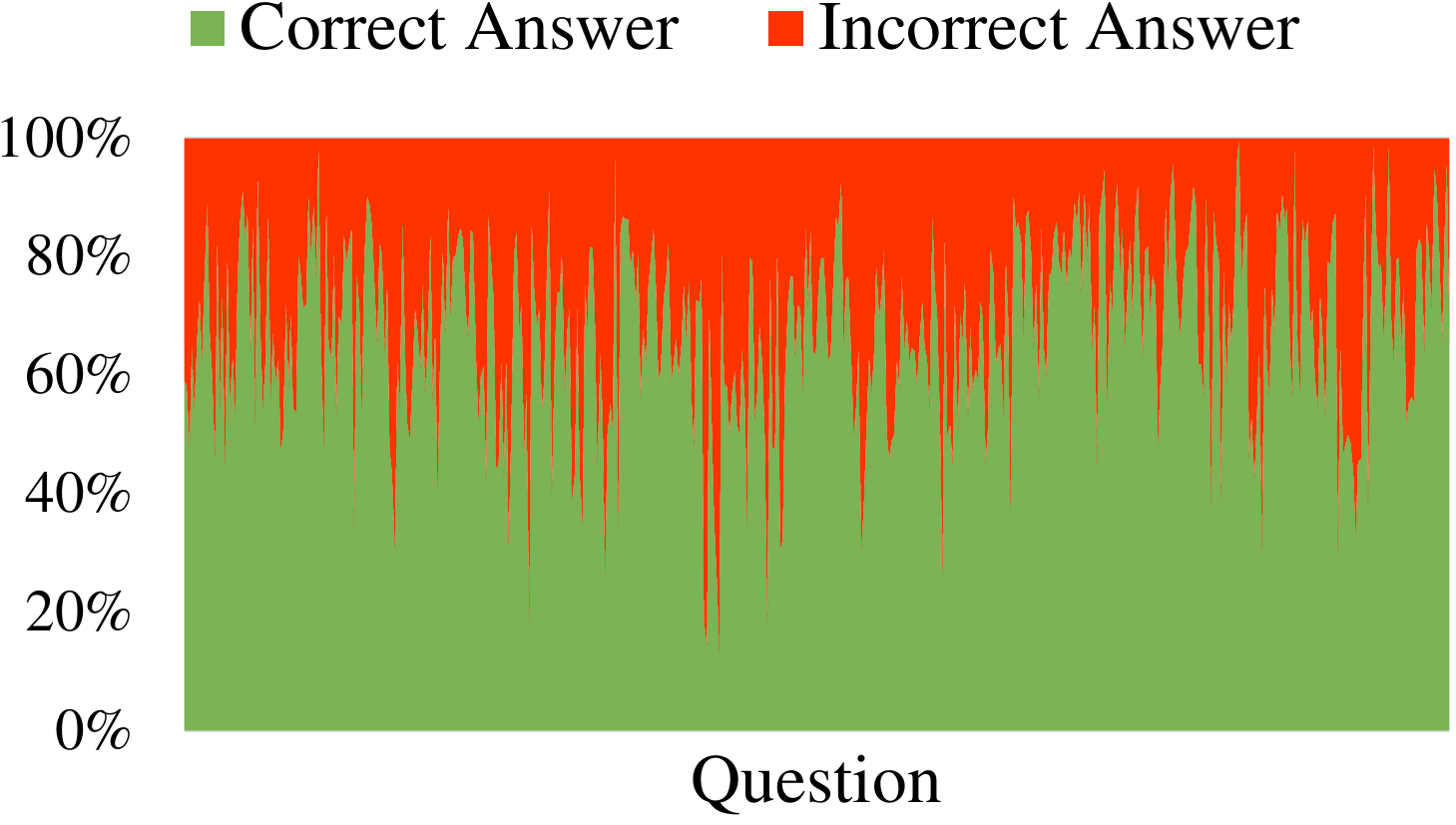}}
    \hspace{0.005\textwidth}
    \subfloat[ASSIST17]
    {\includegraphics[width=0.325\textwidth]{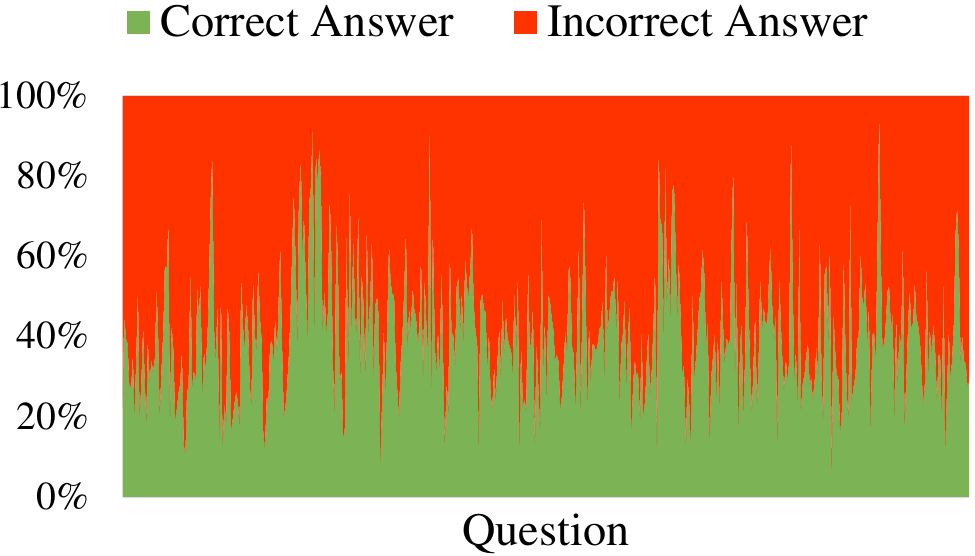}}
    \hspace{0.005\textwidth}
    \subfloat[EdNet]
    {\includegraphics[width=0.325\textwidth]{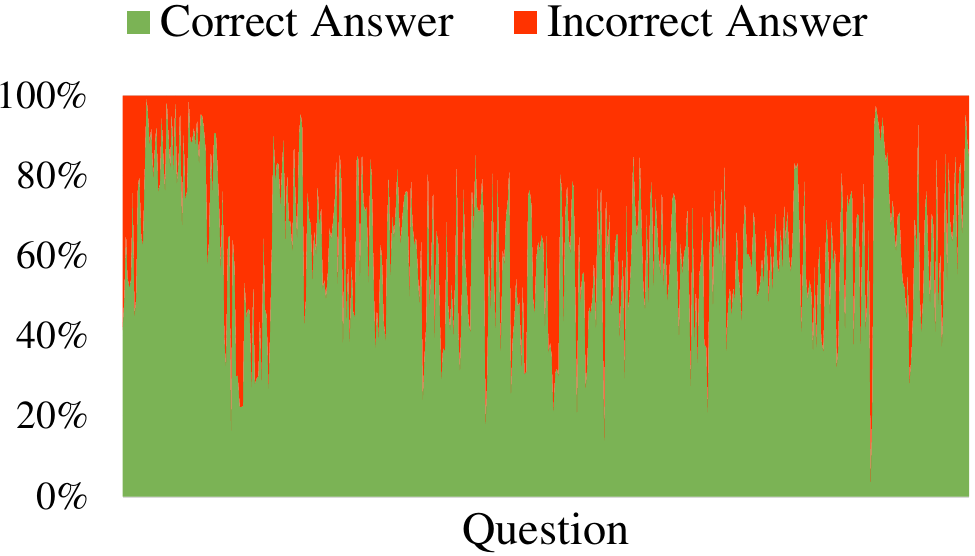}}
     \caption{A percentage stacked bar chart showing the distribution of correct and incorrect answers per question on three benchmark datasets for KT. For display convenience, we randomly selected 500 questions that received at least 20 responses on each dataset.}
    \label{fig:bias_all}
\end{figure}

%Because it is difficult to dynamically obtain the true knowledge states of students, the performance of KT models is indirectly evaluated by predicting how students will perform on future exercises~\cite{liu2021survey}.
%The assumption is that only by understanding students' knowledge states can we estimate their future responses.
It is worth noting that the actual knowledge states of students during the learning process cannot be directly observed.
Therefore, the performance of KT models needs to be indirectly evaluated by predicting how students will perform on future questions~\cite{liu2021survey}.
This is based on the underlying assumption that only by comprehending students' knowledge states can we estimate their future responses.
In general, KT models~\cite{piech2015deep,zhang2017dynamic,ghosh2020context} simplify the student-question interactions by assuming that students' responses are binary, that is, they can either provide a correct or incorrect answer.
%However, we notice that there exists a strong \emph{answer bias} for knowledge concepts, i.e., it is highly unbalanced that the number of correct and incorrect answers to the exercises related to a concept.
%The answer bias can be naturally explained in terms of the difficulty differences between knowledge concepts~\cite{Shen2022Assessing}, but it meanwhile provides us a \emph{shortcut} to achieve good prediction performance in KT.
However, we have observed a strong \emph{answer bias of questions}, suggesting that there is often a significant imbalance between the number of correct and incorrect answers for each question.
In Fig.~\ref{fig:bias_all}, we plot the percentage stacked area charts of correct and incorrect answers across various questions on three benchmark datasets for KT.
The charts reveal that most questions exhibit significantly imbalanced proportions.
The answer bias can be naturally explained in terms of the differences in difficulty between questions~\cite{Shen2022Assessing}.
\emph{Nevertheless, it also provides a shortcut for achieving good prediction performance in KT.}
%Figure~\ref{fig:bias_2009} plots the percentage stacked area chart of the correct and incorrect answers across different concepts on the popular ASSIST09 dataset for KT, from which we can observe that more than two-thirds of concepts have a significantly unbalanced proportion.
%On ASSIST09, if we blindly choose the answer that occurs more often for each concept, a decent accuracy of 67\% can be obtained in testing.
%Furthermore, a simple multi-layer perceptron that uses the concept alone can reach over 70\% accuracy, without even looking at any students' past learning interactions!
After a preliminary analysis, we found that blindly selecting the answer that occurs more frequently for each question yields a decent accuracy in testing, even without taking into account any of the students' past learning interactions!
% Remarkably, this extremely simple rule even surpasses the previous state-of-the-art methods for KT on two datasets, without taking into account any students' past learning interactions!
% Note that these results were obtained without taking into account students' past learning interactions.

In the face of such surprising results, we inevitably raise the issue: \emph{Do existing KT models heavily leverage the answer bias of questions to yield good prediction results?}
To address the issue, we establish a new unbiased evaluation on top of benchmark datasets for KT, in which all students' question-answering interactions in the test set are resampled in such a way that the distribution of correct and incorrect answers per question is balanced (see details in Section~\ref{sec: unbiased_test}).
Fig.~\ref{fig:balancing_all} displays the accuracy of DKT, DKVMN, and AKT on both the original biased test set and the new unbiased test set of different datasets.
Our key finding is that the performance of all methods drops significantly when evaluated on the new unbiased test sets.
% The finding implies that the success of these state-of-the-art KT models is largely attributed to \emph{their ability to memorize the severe answer bias of questions in the dataset}.
% This is problematic because these models give us \emph{a false impression} that they are making progress towards the real goal of understanding students' knowledge states in KT, when they are only exploiting the answer bias to achieve high prediction performance.
The finding suggests that the effectiveness of these state-of-the-art KT models largely stems from \emph{their capacity to memorize the strong answer bias of questions in the dataset}.
This poses a significant problem because these models may create \emph{a misleading impression} that they are advancing towards the goal of understanding students' knowledge states in KT, whereas in reality, they are merely exploiting the answer bias to achieve a high prediction accuracy.

One straightforward solution to mitigate answer bias is to balance the training set by resampling as well.
However, resampling techniques may reduce the sample diversity for training and make models prone to overfitting.
% cause the loss of data information for model learning.
In fact, the answer bias is hard to avoid when collecting real datasets for KT, and our focus should be \emph{how to build unbiased KT models under biased training}.
In this paper, we approach the KT task from a causality perspective~\cite{pearl2000models,pearl2018book}, which has not received sufficient attention in prior research.
A causal graph~\cite{pearl2009causality} is first constructed to formulate the cause-effect relationships among the fundamental components in KT: a student $S$, a question $Q$, the student's knowledge state $K$, and the student's response $R$.
From the causal graph as shown in Fig.~\ref{fig:causal_graph}\subref{subfig: SCM}, we identify that the impact of answer bias lies in the direct causal effect of $Q$ on $R$ and can be mitigated by subtracting this direct effect from the total causal effect.
To this end, we propose a novel COunterfactual REasoning (CORE) framework~\cite{niu2021counterfactual,wei2021model} for KT, which introduces two scenarios, i.e., the conventional KT and counterfactual KT, to estimate the total causal effect and direct causal effect, respectively.
The two scenarios can be described as follows:
\begin{itemize}
  % \item \textbf{Conventional KT}: Whether a student's response $R$ to a concept-related exercise will be correct or not, if models recognize the student $S$ with her/his past learning sequence, see the concept $C$, and understand the knowledge state $K$ of $S$ regarding $C$?
  \item \textbf{In the conventional KT scenario}, a KT model predicts whether a student's response $R$ to a question $Q$ will be correct or not, if the model recognizes the student $S$ based on their past learning sequences, sees the question $Q$, and understands the knowledge state $K$ of $S$ regarding $Q$;
  % \item \textbf{Counterfactual KT}: What $R$ would be, if models only see $C$, but had neither obtained $S$ nor understood $K$?
  \item \textbf{In the counterfactual KT scenario}, a KT model predicts what the student's response $R$ would be, if the model only had access to $Q$, but had neither obtained $S$ nor understood $K$.
\end{itemize}
Intuitively, the conventional KT reflects the actual scenario in existing KT studies, where both $S$ and $Q$ are available, and $K$ can be modeled by matching $S$ and $Q$~\cite{zhang2017dynamic}.
In this case, we can estimate the total causal effect on $R$.
On the other hand, the counterfactual KT involves an imaginary scenario, where $S$ and $K$ had not been accessible, and $R$ would be determined solely by $Q$.
Therefore, the counterfactual KT allows for the estimation of the direct causal effect of $Q$ on $R$, which represents the pure impact of answer bias.

Our CORE framework is model-agnostic, and we instantiate it based upon DKT~\cite{piech2015deep}, DKVMN~\cite{zhang2017dynamic}, and AKT~\cite{ghosh2020context} models, respectively.
During training, an ensemble model is developed to simulate the total causal effect, which consists of an existing KT model and additional network branches capturing the direct causal effects.
The training is performed with a multi-task learning objective, which provides supervision to the learning of different causal effects simultaneously~\cite{niu2021counterfactual,wei2021model}.
In testing, CORE uses the debiased causal effect for inference, which is obtained by subtracting the direct effect related to answer bias from the total effect.
Experimental results on three benchmark datasets demonstrate that CORE outperforms the state-of-the-art methods by large margins in the unbiased evaluation for KT.
% while remaining stable in the biased test setting of different datasets.

In a nutshell, this work makes three main contributions:
\begin{itemize}
  \item We introduce an unbiased evaluation for KT and reveal that the success of many existing models is largely driven by memorizing the answer bias of questions present in datasets, rather than by fully understanding students' knowledge states.
  \item We identify the impact of answer bias as the direct causal effect of questions on students' responses from a causality perspective, and propose a model-agnostic counterfactual reasoning framework CORE to eliminate the impact by making the debiased causal inference for KT.
  \item We instantiate CORE based upon different prevailing KT models and demonstrate its effectiveness and rationality on three benchmark datasets.
\end{itemize}

The remainder of the paper is structured as follows:
Section~\ref{sec: related_work} reviews the related work.
Section~\ref{sec: answer_bias} provides a detailed analysis of the phenomenon of answer bias in KT.
Section~\ref{sec:CORE} presents our counterfactual reasoning framework that reduces the impact of answer bias for KT.
Experimental results and analysis are reported in Section~\ref{sec:experiments}, followed by the conclusions in Section~\ref{sec:conclusions}.

\section{Related Work}\label{sec: related_work}
In this section, we first review the existing literature on KT.
Then, we provide a brief overview of causal inference, which is closely related to our work.

\subsection{Knowledge Tracing}
As the pioneering work of KT, BKT~\cite{corbett1994knowledge} was proposed to utilize a hidden Markov model to represent students' knowledge states as binary variables, indicating whether a knowledge concept is mastered or not.
Furthermore, BKT has been extended to incorporate other important factors, such as students' individual variation in learning~\cite{khajah2016deep} and the prerequisite hierarchies among knowledge concepts~\cite{kaser2017dynamic}.
Pavlik et al.~\cite{Pavlik2009Performance} took a different approach by directly using logistic functions to predict the probability of correctly answering questions according to the side information about questions and students.
Another traditional method of KT is the knowledge tracing machine (KTM) proposed by Vie et al.~\cite{vie2019knowledge}. KTM is based on factorization machines and generalizes logistic models by capturing pairwise feature interactions.
% In spite of the initial breakthrough, these traditional methods adopted relatively simple linear models, which inevitably impeded the progress of knowledge tracing.

With the rapid development of deep learning, various types of neural networks have been investigated to model the learning sequences of students for KT.
Early representative examples include DKT~\cite{piech2015deep} and DKVMN~\cite{zhang2017dynamic}, which are based on RNN and memory network, respectively.
Due to the popularity of the attention mechanism~\cite{vaswani2017attention},  numerous self-attentive models such as SAKT~\cite{pandey2019self}, AKT~\cite{ghosh2020context}, and SAINT~\cite{choi2020towards} have been successively proposed in recent years.
In addition, some works considered more elaborate factors related to learning, including the text content of questions~\cite{liu2019ekt}, the temporal dynamics of different cross-skill impacts~\cite{wang2021temporal,su2021time}, the difficulty levels of questions~\cite{Shen2022Assessing}, and the inter-student information~\cite{long2022improving}.
Liu et al.~\cite{liu2023enhancing} enhanced KT with auxiliary learning tasks, namely the question tagging prediction and individualized prior knowledge prediction.

In another research line, incorporating the relationship information between questions or students is a common practice in graph-based deep learning methods for KT~\cite{song2022bi}.
For example, Nakagawa et al.~\cite{nakagawa2019graph} presented the graph-based knowledge tracing (GKT), which establishes a graph connecting knowledge concepts and models students' knowledge states using graph convolutional networks~\cite{Kipf2017semi}.
Tong et al.~\cite{tong2020structure} proposed the structure-based knowledge tracing (SKT) to discover the knowledge structure of questions and consider the spatial effect on the knowledge structure.
Yang et al.~\cite{yang2021gikt} developed a graph-based interaction model for knowledge tracing (GIKT), which introduces high-order question-concept correlations through embedding propagation.

Despite their impressive performance, existing KT models are overly dependent on the answer bias of questions.
This dependency seriously limits their generalization ability and impedes their progress in achieving the ultimate objective of understanding students' knowledge states in KT.

% \subsection{Causality Analysis}
\subsection{Causal Inference}
% Causality analysis~\cite{pearl2000models,pearl2018book} aims to explore the cause-effect relationships between variables.
Causal inference~\cite{pearl2000models,pearl2018book} aims to explore the cause-effect relationships between variables.
It serves not only as an interpret framework but also as a tool to achieve desired objectives by modeling causal effects.
Recently, causal inference has attracted increasing attention from the machine learning community and has been used in a wide range of applications.
For example, it has been extensively investigated in recommender systems to address issues such as data bias, data missing, and data noise~\cite{zhang2021causerec,zhang2021counterfactual}.
In this study, we follow Pearl's structural causal model~\cite{pearl2009causality} to hypothesize how a student's response is generated in KT through a causal graph.

Counterfactual reasoning is an important concept in causal inference that describes the human introspection behaviors~\cite{wang2021counterfactual}.
In practice, counterfactual reasoning estimates what the situation would be if the treatment variable had a different value from the observed value in the real world.
One important application of counterfactual reasoning is to augment the training samples for data-scarce tasks~\cite{fu2020counterfactual}.
Besides, many researchers exploit counterfactual reasoning to develop more robust, explainable, and fair models in numerous domains, including natural language processing~\cite{li2022learning}, computer vision~\cite{niu2021counterfactual}, and data mining~\cite{wei2021model}.
Our work also falls into this category, and we leverage counterfactual reasoning to mitigate answer bias in KT.
To the best of our knowledge, this is the first attempt to introduce counterfactual reasoning to KT.

\section{Answer Bias in Knowledge Tracing}\label{sec: answer_bias}
In this section, we first formulate the problem of knowledge tracing.
Then, we uncover the prevalence of answer bias of questions across different datasets for KT.
Finally, we establish an unbiased evaluation demonstrating that existing KT methods tend to memorize the answer bias of questions for high prediction performance.

\subsection{Problem Definition}
% In an online learning environment, there is a set of knowledge concepts $\mathcal{C}$.
% All students are asked to answer a series of concept-related exercises to better acquire knowledge.
In an online learning environment, students are required to answer a series of concept-related questions to facilitate their knowledge acquisition.
A $t$-length learning sequence of a student can be represented as $\mathcal{S} = \left[ {\left( {{q_1},{r_1}} \right),\left( {{q_2},{r_2}} \right), \ldots ,\left( {{q_t},{r_t}} \right)} \right]$, where $\left( {{q_l},{r_l}} \right) \in \mathcal{S}$ indicates that the student answers the question $q_l$ at step $l \le t$.
Each question is related to one or more knowledge concepts, and we denote by $\mathcal{C}_{q_l}$ the set of concepts associated with $q_l$.
Consistent with most previous studies~\cite{Abdelrahman2022Knowledge,liu2021survey}, we simplify the student's response as a binary value, i.e., $r_l \in \left\{ {1,0} \right\}$ represents the correctness of the student's answer to $q_l$.

Given the student's learning sequence, KT aims to dynamically assess the student's knowledge state so as to predict their future performance.
A common way to evaluate KT models is by predicting the probability that the student will correctly answer the next question $q$ related to the concepts in $\mathcal{C}_{q}$, i.e., $p\left( {{r} = 1\left| {{\mathcal{S}},{q}, {\mathcal{C}_{q}}} \right.} \right)$.

\begin{table}[t]
\centering
\caption{Performance of a simple baseline that selects the answer appearing more often for KT on different datasets.}
\begin{tabular}{l c c c}
    \toprule
     & ASSIST09 & ASSIST17 & EdNet \\
    \midrule
    Accuracy\rule{0mm}{3mm} & 66.67\% & 66.68\% & 69.25\% \\
    AUC\rule{0mm}{3mm} & 68.41\% & 67.78\% & 70.24\% \\
    \bottomrule
\end{tabular}
\label{tbl:baseline_performance}
\end{table}

\subsection{Answer Bias of Questions}
In this paper, we observe a prevalent phenomenon of answer bias in KT, where the distribution of correct and incorrect answers for questions is often significantly skewed.
To illustrate this phenomenon, Fig.~\ref{fig:bias_all} presents a percentage stacked bar chart showing the distribution of correct and incorrect answers per question on three benchmark datasets for KT, namely ASSIST09, ASSIST17, and EdNet.
Details of these datasets will be presented in Section~\ref{sec:dataset}.
% namely ASSIST09\footnote{https://sites.google.com/site/assistmentsdata/home/2009-2010-assistment-data}, ASSIST17\footnote{https://sites.google.com/view/assistmentsdatamining/dataset?authuser=0}, and Ednet~\cite{choi2020ednet}.
%, and NIPS34~\cite{wang2020diagnostic}.
% In this paper, we observe a prevalent phenomenon of answer bias in KT, where the distribution of correct and incorrect answers for questions is often significantly skewed.
% To quantify this phenomenon, we define the bias strength of a question as the ratio of the more frequent answer type (correct or incorrect) to the total number of answers in the training set.
% It is evident that answer bias is present on all datasets with most questions having a highly uneven proportion of correct and incorrect answers.
As can be seen, the proportion of correct and incorrect answers differs greatly across questions, and all datasets exhibit answer bias with most questions having a highly uneven proportion.

Notably, answer bias can lead to seemingly impressive performance for KT.
For instance, we can blindly predict students' responses as the answer that appears more frequently during training for a given question.
Table~\ref{tbl:baseline_performance} lists the accuracy and area under ROC curve (AUC) results of this simple baseline on different datasets.
It proves to be competitive, even though it does not consider any of the students' past learning interactions.
These findings confirm our belief that leveraging the answer bias of questions is a shortcut to achieve good performance in KT.

%\begin{figure*}[t]
%    \centering
%    \subfloat[ASSIST09]
%    {\includegraphics[width=0.4\textwidth]{pic/balancing 09.pdf}}
%    \hspace{0.05\textwidth}
%    \subfloat[ASSIST17]
%    {\includegraphics[width=0.4\textwidth]{pic/balancing 17.pdf}}
%    \\
%    \subfloat[EdNet]
%    {\includegraphics[width=0.4\textwidth]{pic/balancing EdNet.pdf}}
%    \hspace{0.05\textwidth}
%    \subfloat[NIPS34]
%    {\includegraphics[width=0.4\textwidth]{pic/balancing NIPS34.pdf}}
%     \caption{Performance changes of DKT, DKVMN, AKT between the original biased test set and the new unbiased test set of different datasets in terms of accuracy and AUC, respectively.}
%    \label{fig:balancing_all}
%\end{figure*}

\begin{figure}[t]
    \centering
    \subfloat[ASSIST09]
    {\includegraphics[width=0.325\textwidth]{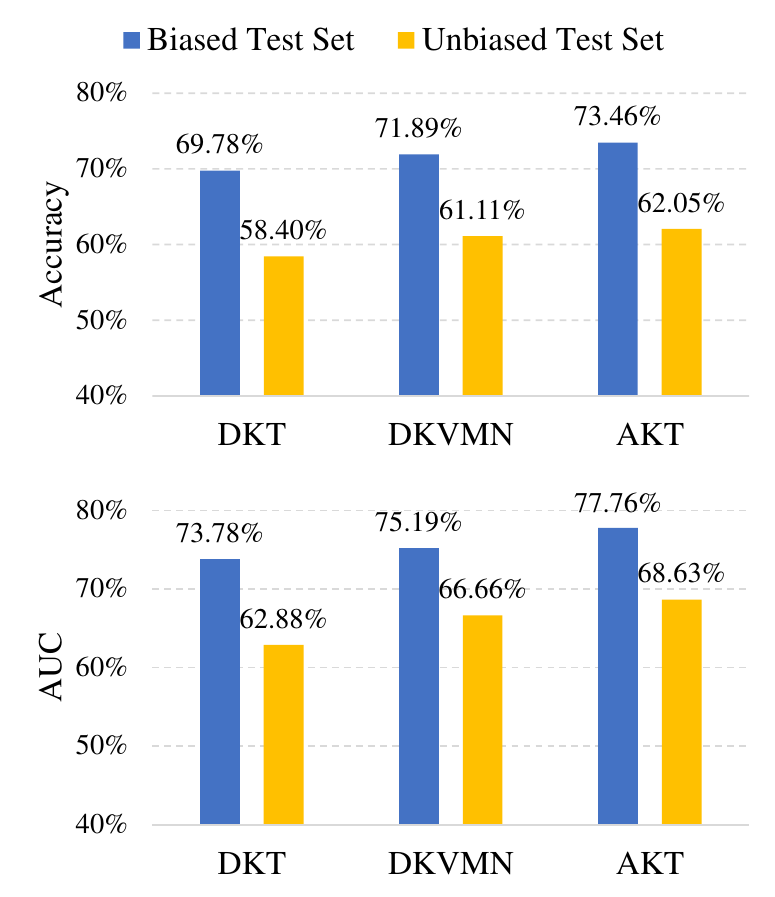}}
    \hspace{0.005\textwidth}
    \subfloat[ASSIST17]
    {\includegraphics[width=0.325\textwidth]{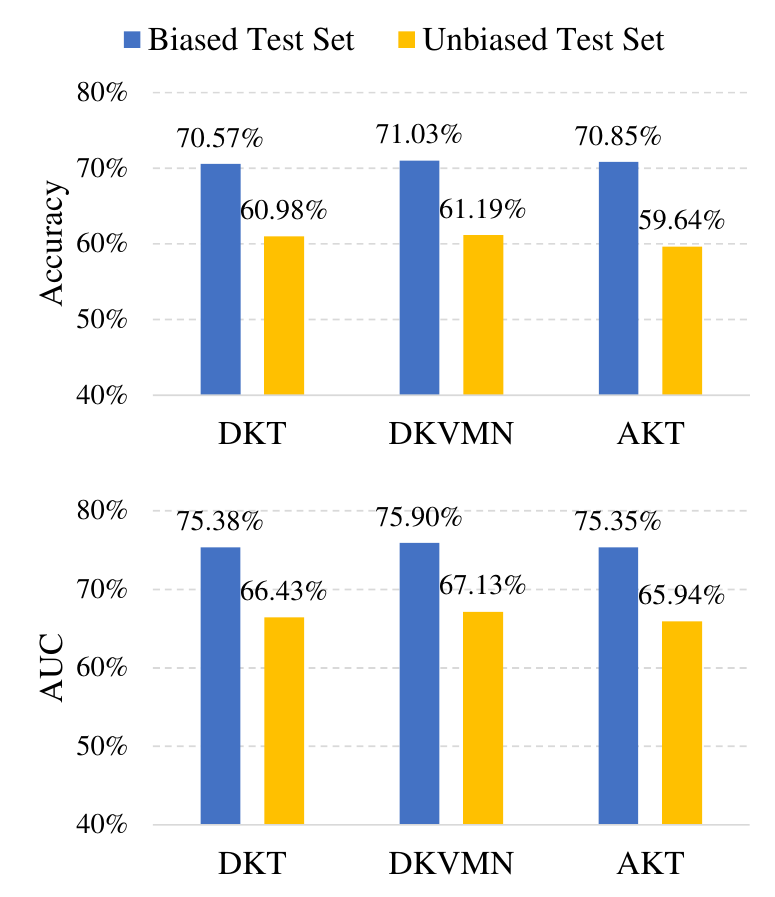}}
    \hspace{0.005\textwidth}
    \subfloat[EdNet]
    {\includegraphics[width=0.325\textwidth]{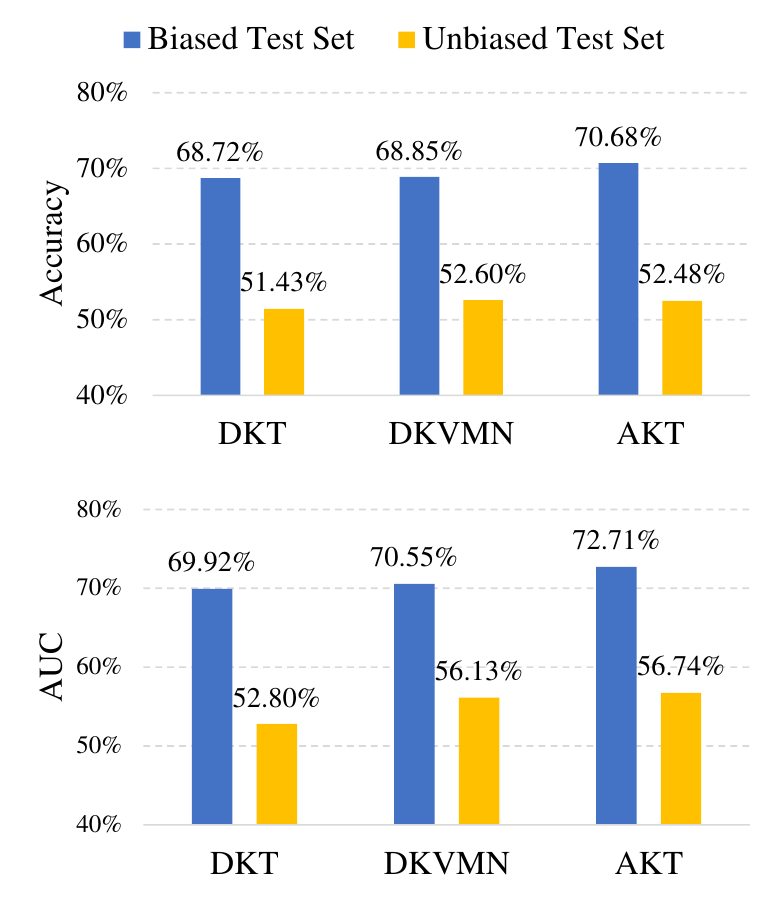}}
     \caption{Performance changes of DKT, DKVMN, AKT between the original biased test set and the new unbiased test set of different datasets in terms of accuracy and AUC, respectively.}
    \label{fig:balancing_all}
\end{figure}

\subsection{Unbiased Evaluation}\label{sec: unbiased_test}
In recent years, we have witnessed the emergence of increasingly sophisticated models that achieve higher and higher performance for KT.
However, the existence of answer bias makes it complicated to determine the source of these improvements.
It is unclear whether the models have fully understood students' knowledge states or whether their performance is heavily attributed to memorizing answer bias inherent in the training data.

To disentangle the influence of the two factors, we set up a new unbiased evaluation on top of benchmark datasets for KT.
Specifically, we randomly sampled learning interactions with each question for all students in the test set, achieving a balance between correct and incorrect answers.
During this process, some interactions may be sampled more than once, while others may be excluded.
To maintain the testing scale, we ensure that the number of interactions with each question after sampling is equivalent to that in the original test set.
\emph{Note that we only resampled the test set and left the training set unchanged.
In testing, a trained model assesses students' knowledge states by analyzing their complete sequences of past learning interactions, while the model's predictions are only evaluated based on the sampled interactions.
This is equivalent to configuring the model to predict only for a subset of test interactions with balanced correct and incorrect answers.}
Intuitively, our unbiased evaluation requires KT models to rely on a deep understanding of students' knowledge states to achieve high performance, instead of simply memorizing the answer bias to game the test.
After all, with balanced correct and incorrect answers, it is insufficient to blindly assume the answer without considering students' past learning interactions.
Therefore, we believe that our unbiased evaluation more accurately reflects the progress in knowledge state understanding for KT.

Fig.~\ref{fig:balancing_all} shows the changes in performance of DKT, DKVMN, AKT between the original biased test set and the new unbiased test set of different datasets in terms of accuracy and AUC, respectively.
Unfortunately, all methods experience a sharp degradation in performance on the new unbiased test set compared to the original biased test set.
% For example, the average drop in accuracy across all cases is more than 10\%.
More precisely, the average accuracy decreases by 11.19\%, 10.21\%, and 17.25\%, and the average AUC declines by 9.52\%, 9.04\%, and 15.84\% on ASSIST09, ASSIST17, and EdNet, respectively.
These results demonstrate that the success of these models largely stems from memorizing the severe answer bias present in the datasets.
When the answer bias is removed in the new test set, the models' advantage disappears.
We argue that this is problematic because existing KT models appear to perform well in predicting students' future responses, but in reality, they fail to fully understand students' knowledge states, which is the ultimate goal of KT.

\section{Causality Perspective of Knowledge Tracing}~\label{sec:CORE}
In this paper, we explore the KT task from a causality perspective~\cite{pearl2000models,pearl2018book}.
Firstly, we build a causal graph of KT and recognize the impact of answer bias as the direct causal effect of questions on students' responses.
Next, we present a COunterfactual REasoning (CORE) framework to overcome answer bias by removing this direct causal effect.
Finally, we provide the implementations of CORE based on DKT, DKVMN, and AKT models, respectively.

\begin{figure}[t]
    \centering
    \subfloat[Causal Graph]
    {\label{subfig: SCM}
    \includegraphics[width=0.22\textwidth]{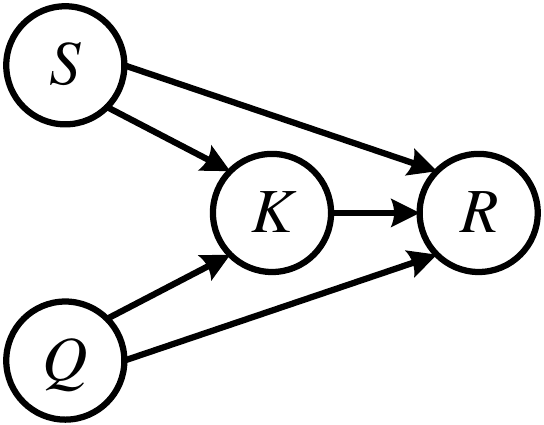}
    }
    \hspace{0.05\textwidth}
    \subfloat[Difference between TE and NDE in Counterfactual Reasoning]
    {\label{subfig: TIE}
    \includegraphics[width=0.5\textwidth]{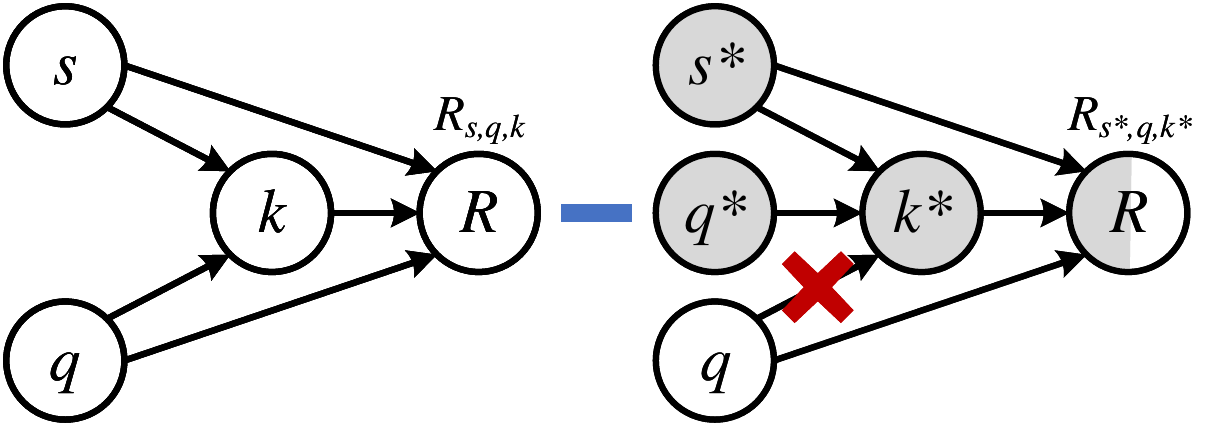}
    }
    \caption{(a) The causal graph of KT. (b) The difference between TE in the conventional KT and NDE in the counterfactual KT. White nodes are at the value $S=s$ and $Q=q$, while gray nodes are at the value $S=s^\ast$ and $Q=q^\ast$.}
    \label{fig:causal_graph}
\end{figure}

\subsection{Causal Graph}
Generally, KT is considered as a binary classification task, i.e., predicting whether a student will give a correct response to a concept-related question or not.
We build a causal graph~\cite{pearl2009causality} to model the cause-effect relationships among the fundamental components in KT: a student $S$, a question $Q$, the student's knowledge state $K$, and the student's response $R$.
The causal graph is represented by a directed acyclic graph, where each node corresponds to a variable, and each edge signifies a relationship among variables.
As illustrated in Fig.~\ref{fig:causal_graph}\subref{subfig: SCM}, the causal graph of KT is established by linking the variables as follows:
\begin{itemize}
  \item \bm{$S,Q \to K \to R$} represents the indirect causal effect of $S$ and $Q$ on $R$ via $K$, and the majority of existing KT models can be interpreted along this causal chain.
      Specifically, the models first encode the question representation $Q=q$, and then acquire the student representation $S=s$ by analyzing the student's historical learning sequence.
      $s$ and $q$ are matched to estimate the knowledge state of $s$ regarding $q$~\cite{zhang2017dynamic}, i.e., $k = K_{s,q} = K\left({S=s, Q=q}\right)$, where $K\left( \cdot \right)$ denotes the value function of $K$.
      Finally, $k$ is utilized to predict the value of $R$.
  \item \bm{$S \to R$} reflects the direct causal effect of $S$ on $R$. As noted in previous studies~\cite{khajah2016deep,long2021tracing}, students exhibit diverse study habits and cognition abilities when tackling the same question.
      For various reasons, students may correctly guess the answers to unfamiliar questions while making errors on the questions they have already mastered~\cite{corbett1994knowledge}.
      Such phenomena can be attributed to this causal path.
  \item \bm{$Q \to R$} captures the direct causal effect of $Q$ on $R$, which has been overlooked in prior research. As discussed earlier, if the answer bias of questions is memorized, KT models can achieve good performance by using the questions alone.
      Therefore, this causal path highlights the impact of answer bias on predicting students' responses.
\end{itemize}

\subsection{Counterfactual Reasoning}
Causal inference involves quantifying the causal effect by comparing two potential outcomes of the same individual under two different treatments~\cite{rubin1978bayesian,robins1986new}.
For example, the effect of a drug on a disease can be measured by the difference in health state of patients who take the drug (i.e., under the treatment condition) versus those who do not take the drug (i.e., under the no-treatment condition).

Given the three paths connected to $R$ in Fig.~\ref{fig:causal_graph}\subref{subfig: SCM}, i.e., $S \to R$, $Q \to R$, and $K \to R$, the value of $R$ can be determined by the values of $S$ and $Q$, i.e.,
\begin{equation}\label{eq:actual_function}
  R_{s,q,k} = R\left( {S=s, Q=q, K=k} \right),
\end{equation}
where $R\left( \cdot \right)$ denotes the value function of $R$.
Note that the value of $K$ can be obtained from $k = K_{s,q} = K\left({S=s, Q=q}\right)$.
In the conventional KT scenario, the total effect (TE) of $S=s$ and $Q=q$ on $R$ is defined as the difference between the potential outcomes under the treatment condition $S=s$, $Q=q$, and $K=k$ and under the no-treatment condition $S=s^\ast$, $Q=q^\ast$, and $K=k^\ast$:
\begin{equation}\label{eq:TE}
  TE = {R_{s,q,k}} - {R_{s^\ast,q^\ast,k^\ast}},
\end{equation}
where $s^\ast$ and $q^\ast$ refer to the situation when the values of $S$ and $Q$ are unknown and are typically set as null.
$k^\ast = K_{s^\ast,q^\ast}$ represents the value of $K$ when $S=s^\ast$ and $Q=q^\ast$.

In this paper, we argue that existing KT models are heavily dependent on the answer bias of questions, which limits their ability to infer students' responses by fully understanding their knowledge states.
To address this issue, we propose a COunterfactual REasoning (CORE) framework to improve KT models by mitigating the impact of answer bias.
From the causal graph of KT in Fig.~\ref{fig:causal_graph}\subref{subfig: SCM}, we identify that the direct causal effect along $Q \to R$ explicitly reflects the impact of answer bias, which can be estimated in a counterfactual KT scenario.
The counterfactual KT depicts an imaginary scenario, where only the question is accessible, and neither the student nor the knowledge state is given.
In this scenario, $Q$ is set to $q$, but $S$ is set to $s^\ast$, and $K$ takes the value $k^\ast$ when $S = s^\ast$ and $Q = q^\ast$:
\begin{equation}\label{eq:counterfactual_function}
  R_{s^\ast,q,k^\ast} = R\left( {S=s^\ast, Q=q, K=k^\ast} \right).
\end{equation}
It is important to note that only in the counterfactual world can $Q$ be set to different values at the same time, i.e., $Q = q$ and $Q = q^\ast$ for calculating $R_{s^\ast,q,k^\ast}$ and $k^\ast = K_{s^\ast,q^\ast}$, respectively.
In Eq.~\eqref{eq:counterfactual_function}, since the values of $S$ and $K$ are both ignored, KT models have to rely solely on the given question $Q=q$ for decision making.
Under this setting, the natural direct effect (NDE) of $Q$ on $R$ can be defined as the decrease in $R$ when $Q$ changes from the treatment condition $q$ to the no-treatment condition $q^\ast$:
\begin{equation}\label{eq:NDE}
  NDE = {R_{s^\ast,q,k^\ast}} - {R_{s^\ast,q^\ast,k^\ast}}.
\end{equation}

To mitigate the impact of answer bias, it is necessary to subtract the direct effect of $Q$ on $R$ from the total effect, i.e.,
\begin{equation}\label{eq:TIE}
  TE - NDE = {R_{s,q,k}} - {R_{s^\ast,q,k^\ast}}.
\end{equation}
Fig.~\ref{fig:causal_graph}\subref{subfig: TIE} shows the difference between $TE$ and $NDE$, which is the debiased causal effect on $R$ and is used for inference in our CORE framework.

\subsection{Model Implementation}\label{sec: model_implementation}
\begin{figure}
    \centering
    \includegraphics[width=0.5\textwidth]{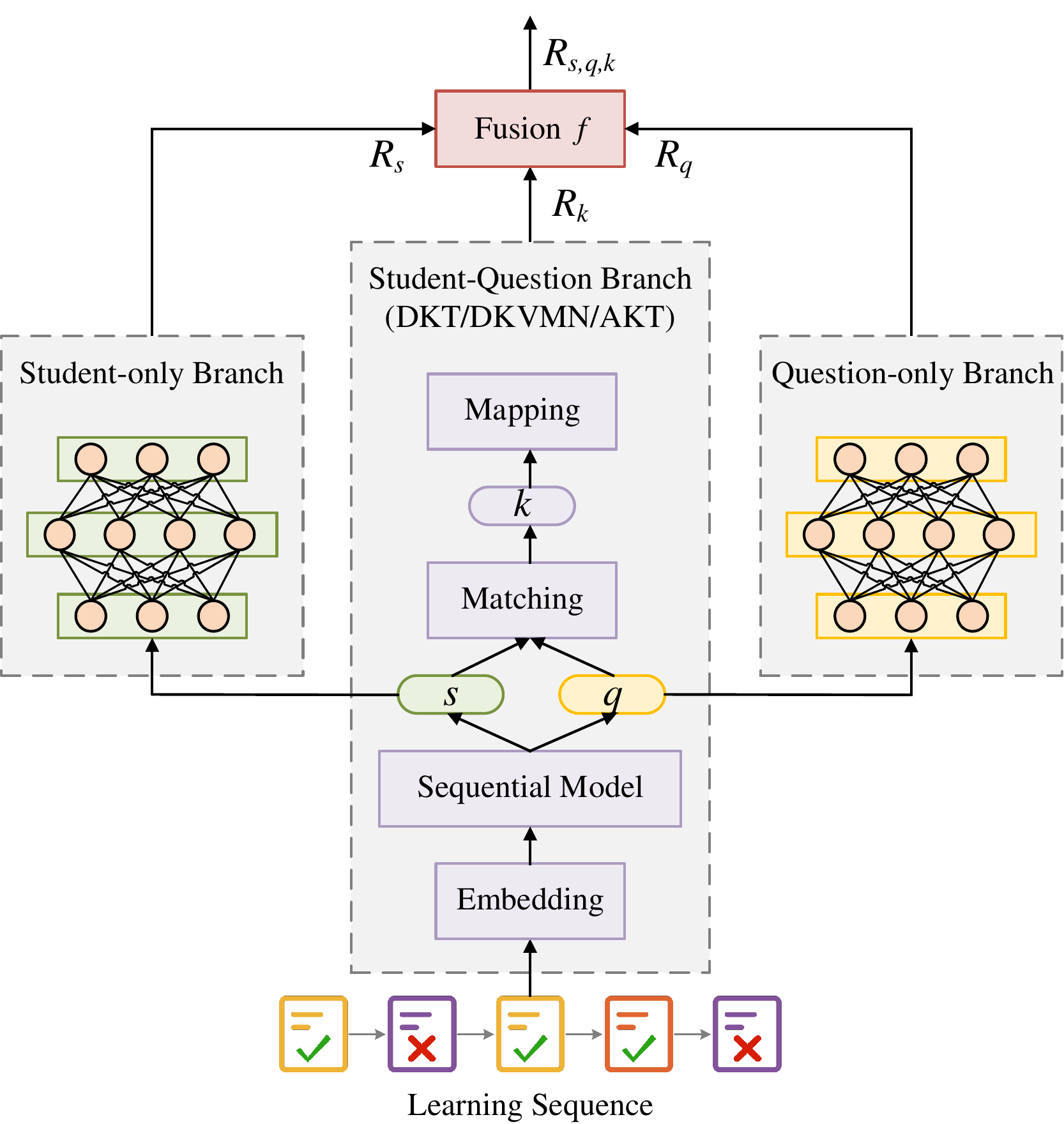}
    \caption{The conceptual structure of the CORE framework. It consists of three network branches that capture each causal path in the causal graph of KT, respectively.}
    \label{fig:architecture}
\end{figure}

Our CORE framework is model-agnostic, meaning that it can be implemented with different existing KT models.
The conceptual structure of the CORE framework is depicted in Fig~\ref{fig:architecture}.

\subsubsection{Parameterization}
Eq.~\eqref{eq:actual_function} shows that the calculation of the response $R_{s,q,k}$ involves the values $s$, $q$, and $k$.
These values can be parameterized by three neural network branches and a fusion function as follows:
\begin{equation}\label{eq:actual_parameterization}
\begin{array}{c}
{R_s} = {\mathcal{B}_S}\left( s \right),\ {R_q} = {\mathcal{B}_Q}\left( q \right),\ {R_k} = {\mathcal{B}_{SQ}}\left( {s,q} \right),\\
{R_{s,q,k}} = f\left( {{R_s},{R_q},{R_k}} \right).
\end{array}
\end{equation}
The calculation is in line with the causal graph of KT, where ${\mathcal{B}_S}$ represents the student-only branch along $S \to R$, ${\mathcal{B}_Q}$ represents the question-only branch along $Q \to R$, and ${\mathcal{B}_{SQ}}$ represents the student-question branch along $S,Q \to K \to R$.
The function $f$ combines the outputs of the three branches to obtain the final response $R_{s,q,k}$.

The response $R_{s^\ast,q,k^\ast}$ needs to be computed in the counterfactual KT scenario, where $S$ is set to the null value $s^\ast$ and $K$ is obtained with the null values $s^\ast$ and $q^\ast$.
However, a neural network cannot accept null values as input, so we expect that the model is able to learn to generate the outputs $R_{s^\ast}$ and $R_{k^\ast}$ of ${\mathcal{B}_S}$ and ${\mathcal{B}_{SQ}}$, respectively.
In this case, we can compute $R_{s^\ast,q,k^\ast}$ as:
\begin{equation}\label{eq:counterfactual_parameterization}
\begin{array}{c}
{R_{s^\ast}} = p,\ {R_q} = {\mathcal{B}_Q}\left( q \right),\ {R_{k^\ast}} = p,\\
{R_{s^\ast,q,k^\ast}} = f\left( {{R_{s^\ast}},{R_q},{R_{k^\ast}}} \right),
\end{array}
\end{equation}
where $p$ denotes a learnable model parameter.
For simplicity, we use the same parameter $p$ for both $R_{s^\ast}$ and $R_{k^\ast}$.

\subsubsection{Instantiation}
As mentioned above, most existing KT models follow the causal chain of $S,Q \to K \to R$.
In our implementation, we select the prevalent DKT~\cite{piech2015deep}, DKVMN~\cite{zhang2017dynamic}, and AKT~\cite{ghosh2020context} models as the backbone of the student-question branch ${\mathcal{B}_{SQ}}$, respectively.
Similar to the recent works~\cite{long2021tracing,chen2023improving}, each question $q$ is represented as a combination of the question's embedding and the embedding of its corresponding concepts, i.e.,
\begin{equation}
q = \bm{e}_q \oplus \overline {{\bm{e}_c}},
\end{equation}
where $\bm{e}_q \in \mathbb{R}^d$ denotes the question's embedding, and $\oplus$ denotes the concatenate operator.
$\overline {{\bm{e}_c}} \in \mathbb{R}^d$ is the average embedding of all knowledge concepts associated with the question, i.e.,
\begin{equation}
\overline {{\bm{e}_c}} = \frac{1}{{\left| {{\mathcal{C}_q}} \right|}}\sum\limits_{c \in {\mathcal{C}_q}} {{\bm{e}_c}}.
\end{equation}
Given the learning sequence of a student $\mathcal{S} = \left\{ {\left( {{q_1},{r_1}} \right),\left( {{q_2},{r_2}} \right), \ldots ,\left( {{q_t},{r_t}} \right)} \right\}$, each learning interaction $\left( {{q_l},{r_l}} \right)$ can be represented by
\begin{equation}
i_l = \left\{ \begin{aligned}
&q_l \oplus \bm{0}, &&r_l = 1;\\
&\bm{0} \oplus q_l, &&r_l = 0.
\end{aligned} \right.
\end{equation}
DKT, DKVMN, and AKT are built based on the sequential models RNN~\cite{hochreiter1997long}, memory network~\cite{sukhbaatar2015end}, and Transformer~\cite{vaswani2017attention}, respectively.
They process the learning interaction $i_l$ at each step and encode the learning sequence $\mathcal{S}$ into the student embedding $s$.
$k$ represents the mastery level of the student $s$ on the question $q$, i.e., the knowledge state of $s$ regarding $q$, which is generally computed based on the matching degree between $s$ and $q$.
Finally, $k$ is mapped to the output $R_k$ of ${\mathcal{B}_{SC}}$.
Here, we omit the technical details of DKT, DKVMN, and AKT, which can be found in their original papers~\cite{piech2015deep,zhang2017dynamic,ghosh2020context}.

For the student-only branch ${\mathcal{B}_{S}}$ and the question-only branch ${\mathcal{B}_{Q}}$, they are simply designed as two multi-layer perceptrons that project the student representation $s$ and the question representation $q$ into the outputs $R_s$ and $R_q$, respectively.
The fusion function $f$ is the sum combination of $R_s$, $R_q$, and $R_k$ followed by non-linear transformations:
\begin{equation}\label{eq: fusion}
{R_{s,q,k}} = f\left( {{R_{s}},{R_q},{R_{k}}} \right) = \log \sigma \left( {{R_s} + {R_q} + {R_k}} \right),
\end{equation}
where $\sigma$ denotes the sigmoid function.
Note that more complicated branch designs and fusion operations can be adopted, but we leave them for future exploration.

\subsubsection{Training}
In our implementation, the outputs of network branches $R_s$, $R_q$, and $R_k$, as well as their combination ${R_{s,q,k}}$, are all formulated as the likelihood of the student $s$ correctly answering the question $q$ at the next step.
To learn the model parameters, we minimize a binary cross-entropy (BCE) loss~\cite{long2022improving} between the predicted probability and the student's true response:
%\begin{equation}\label{eq: loss_k}
%\begin{split}
%{\mathcal{L}_{BCE}^{sq}} =  -& r\log \sigma \left( {{R_{s,q,k}}} \right) \\
%-& \left( {1 - r} \right)\log \left( {1 - \sigma \left( {{R_{s,q,k}}} \right)} \right),
%\end{split}
%\end{equation}
\begin{equation}\label{eq: loss_k}
{\mathcal{L}_{BCE}^{sq}} =  - r\log \sigma \left( {{R_{s,q,k}}} \right) - \left( {1 - r} \right)\log \left( {1 - \sigma \left( {{R_{s,q,k}}} \right)} \right),
\end{equation}
where $r \in \left\{ {1,0} \right\}$ indicates whether the student actually gives the correct response or not.
% To better isolate the direct effects of the student $s$ and question $q$ on the response $r$ from the total effect, we further introduce a multi-task learning strategy~\cite{niu2021counterfactual,wei2021model} that imposes additional BCE losses on $R_s$ and $R_q$:
%\begin{equation}\label{eq: loss_sc}
%\begin{array}{c}
%{\mathcal{L}_{BCE}^{s}} =  - r\log \sigma \left( {{R_{s}}} \right) - \left( {1 - r} \right)\log \left( {1 - \sigma \left( {{R_{s}}} \right)} \right),\\
%{\mathcal{L}_{BCE}^{q}} =  - r\log \sigma \left( {{R_{q}}} \right) - \left( {1 - r} \right)\log \left( {1 - \sigma \left( {{R_{q}}} \right)} \right).
%\end{array}
%\end{equation}
To better model the direct effect of the question $q$ on the student's response, we additionally apply the BCE loss directly to $R_q$:
\begin{equation}\label{eq: loss_q}
{\mathcal{L}_{BCE}^{q}} =  - r\log \sigma \left( {{R_{q}}} \right) - \left( {1 - r} \right)\log \left( {1 - \sigma \left( {{R_{q}}} \right)} \right).
\end{equation}

The above calculations of BCE losses do not involve the model parameter $p$ in Eq.~\eqref{eq:counterfactual_parameterization}.
% To estimate the value of $p$, we introduce a constraint that requires the response $R_{s^\ast,q,k^\ast}$ in the counterfactual KT to mimic the response $R_{s,q,k}$ in the conventional KT.
To estimate the value of $p$, we hypothesize that the distribution of correct and incorrect answers in the counterfactual KT should not deviate much from that in the conventional KT.
Otherwise, an improper $p$ would lead to the debiased causal effect in Eq.~\eqref{eq:TIE} being dominated by TE or NDE~\cite{niu2021counterfactual}.
Therefore, we measure the Kullback-Leibler (KL) divergence between the distributions obtained from $R_{s,q,k}$ and $R_{s^\ast,q,k^\ast}$:
%\begin{equation}\label{eq: loss_kl}
%\begin{split}
%{\mathcal{L}_{KL}} = - & \sigma \left( {{R_{s,q,k}}} \right)\log \frac{{\sigma \left( {{R_{s^\ast,q,k^\ast}}} \right)}}{{\sigma \left( {{R_{s,q,k}}} \right)}}\\
% - & \left( {1 - \sigma \left( {{R_{s,q,k}}} \right)} \right)\log \frac{{1 - \sigma \left( {{R_{s^\ast,q,k^\ast}}} \right)}}{{1 - \sigma \left( {{R_{s,q,k}}} \right)}}.
%\end{split}
%\end{equation}
\begin{equation}\label{eq: loss_kl}
{\mathcal{L}_{KL}} = - \sigma \left( {{R_{s,q,k}}} \right)\log \frac{{\sigma \left( {{R_{s^\ast,q,k^\ast}}} \right)}}{{\sigma \left( {{R_{s,q,k}}} \right)}} - \left( {1 - \sigma \left( {{R_{s,q,k}}} \right)} \right)\log \frac{{1 - \sigma \left( {{R_{s^\ast,q,k^\ast}}} \right)}}{{1 - \sigma \left( {{R_{s,q,k}}} \right)}}.
\end{equation}
Note that only $p$ is updated when minimizing $\mathcal{L}_{KL}$.
Overall, the optimization problem for training the proposed CORE can be solved by iteratively performing the following two steps:
\begin{equation}\label{eq: optimization}
\begin{split}
% \mathop {\min }\limits_{\mathcal{B}_S,\mathcal{B}_Q,\mathcal{B}_{SQ}}\ &{{\mathcal{L}_{BCE}^{sq}} + {\mathcal{L}_{BCE}^{s}} + {\mathcal{L}_{BCE}^{q}}},\\
\mathop {\min }\limits_{\mathcal{B}_S,\mathcal{B}_Q,\mathcal{B}_{SQ}}\ &{{\mathcal{L}_{BCE}^{sq}} + {\mathcal{L}_{BCE}^{q}}},\\
\mathop {\min }\limits_p\ &{{\mathcal{L}_{KL}}}.
\end{split}
\end{equation}

\subsubsection{Inference}
After the training is completed, we use the debiased causal effect for inference, which is implemented as:
%\begin{equation}\label{eq: inference}
%\begin{split}
%TE - NDE =\ & R_{s,q,k} - R_{s^\ast,q,k^\ast} \\
%=\ & f\left( {{R_{s}},{R_q},{R_{k}}} \right) - f\left( {{R_{s^\ast}},{R_q},{R_{k^\ast}}} \right).
%\end{split}
%\end{equation}
\begin{equation}\label{eq: inference}
TE - NDE = R_{s,q,k} - R_{s^\ast,q,k^\ast} = f\left( {{R_{s}},{R_q},{R_{k}}} \right) - f\left( {{R_{s^\ast}},{R_q},{R_{k^\ast}}} \right).
\end{equation}

\section{Experiments}\label{sec:experiments}
In this section, we provide a detailed description of the experimental setup and present a series of experimental results to validate the efficacy of our CORE framework.

\subsection{Dataset}\label{sec:dataset}
Experiments were conducted on three benchmark datasets for KT, namely, ASSIST09\footnote{https://sites.google.com/site/assistmentsdata/home/2009-2010-assistment-data/skill-builder-data-2009-2010}, ASSIST17\footnote{https://sites.google.com/view/assistmentsdatamining/dataset?authuser=0}, and EdNet\footnote{https://github.com/riiid/ednet}~\cite{choi2020ednet}.
ASSIST09 is made up of math exercises collected from the free online tutoring platform ASSISTments during the 2009-2010 school year.
ASSIST17 is derived from the 2017 ASSISTments data mining competition.
EdNet is a very large dataset collected by Santa, a multi-platform AI tutoring service, containing over 130 million learning interactions from approximately 780,000 students.
To ensure computational efficiency, we randomly selected 5,000 students from EdNet as previous works~\cite{liu2020improving,wu2022sgkt} did.
For all datasets, questions without knowledge concepts and students with fewer than three learning interactions were filtered out.
The dataset statistics are summarized in Table~\ref{tab:dataset}.

\begin{table}[t]
\caption{Dataset statistics.}
\centering
\begin{tabular}{l c c c}
    \toprule
     & ASSIST09 & ASSIST17 & EdNet \\
    \midrule
    \# Students\rule{0mm}{3mm} & 3,852 & 1,708 & 4,718 \\
    % \# Exercises & 16,891 & 3,162 & 10,795 \\
    \# Questions\rule{0mm}{3mm} & 17,737 & 3,162 & 11,955 \\
    \# Concepts\rule{0mm}{3mm} & 123 & 102 & 188 \\
    \# Interactions\rule{0mm}{3mm} & 282,619 & 942,814 & 586,952 \\
    \# Interactions per question\rule{0mm}{3mm}  & 15.93 & 298.17 & 49.10 \\

    \bottomrule
\end{tabular}
\label{tab:dataset}
%\end{center}
\end{table}

\subsection{Evaluation Metrics}
As the original test sets of different datasets have severe answer bias, evaluating models on these original test sets fails to reflect their true ability to understand students' knowledge states.
Therefore, we propose to establish the new unbiased test set for each dataset following the procedures outlined in Section~\ref{sec: unbiased_test}.
All models are evaluated through a binary classification task that predicts whether a student will answer the next question correctly or not.
The classification accuracy and AUC values are used as the evaluation metrics.

\subsection{Baselines}
We compared our methods against several well-known KT models, including:
\begin{itemize}
  \item \textbf{BKT}~\cite{corbett1994knowledge} is based on the hidden Markov model and represents students' knowledge states for each concept as a binary variable.
  \item \textbf{DKT}~\cite{piech2015deep} processes the learning sequences of students with RNN and uses the hidden states of RNN to represent the knowledge states of students at each step.
% adopts RNN to process the learning sequences of students and represents their knowledge states with the hidden states of RNN.
  \item \textbf{DKVMN}~\cite{zhang2017dynamic} introduces the memory network to represent knowledge concepts in a key matrix and to store and update the mastery of corresponding concepts of students in a value matrix.
  \item \textbf{AKT}~\cite{ghosh2020context} uses the Rasch model to generate question embeddings and applies the Transformer architecture to model the learning sequences of students. A monotonic attention mechanism is also introduced to reduce the importance of interactions occurred in the distant past.
  \item \textbf{SAKT}~\cite{pandey2019self} uses the self-attention mechanism to identify the past learning interactions related to the question to be answered and further makes the prediction by focusing on these relevant interactions.
  \item \textbf{SAINT}~\cite{choi2020towards} is also built upon Transformer, but it separates the question sequence and the response sequence and feeds them to the encoder and the decoder, respectively.
\end{itemize}
We implemented our CORE framework with DKT, DKVMN, and AKT, leading to their debiased counterparts named \textbf{DKT-CORE}, \textbf{DKVMN-CORE}, and \textbf{AKT-CORE}, respectively.

\subsection{Implementation Details}
On each dataset, 80\% of the students were used for training with 5-fold cross validation, while the remaining 20\% were reserved for testing.
The maximum length of students' learning sequences was set to 200.
If a student has more than 200 question-answering interactions, we partitioned their learning sequence into multiple subsequences.
The dimension of all embeddings and hidden states was set to 64.

For the sake of reproducibility, we adopted the publicly available \textsc{pyKT}\footnote{https://github.com/pykt-team/pykt-toolkit}~\cite{liupykt2022} benchmark implementations of all baseline methods.
The hyperparameters of baseline methods were set to the values specified in their original papers.
Our models were trained using the mini-batch Adam optimizer~\cite{kingma2014adam}.
We configured the batch size to be 128.
The learning rate was $10^{-3}$, and the maximum number of epochs during training was 200.
The Pytorch code of our models has been released at \url{https://github.com/lucky7-code/CORE}.

\subsection{Performance Comparison}
Table~\ref{tbl:unbiased_performance} shows the performance comparison among different methods on the unbiased test sets of different datasets.
Overall, our CORE framework shows excellent robustness when the answer bias is removed in testing.
In Fig.~\ref{fig:balancing_all} presented above, we have illustrated the significant drop in performance for DKT, DKVMN, and AKT when evaluated on the unbiased test sets.
Fortunately, our CORE framework effectively compensates for these performance decreases.
Specifically, DKT-CORE exhibits an average improvement of 4.75\% and 5.25\% in accuracy and AUC compared to DKT across the three datasets.
Similarly, the average improvement of DKVMN-CORE over DKVMN is 2.10\% and 1.50\%, and that of AKT-CORE over AKT is 2.24\% and 0.85\%, respectively.
These impressive results highlight the effectiveness of CORE in mitigating answer bias by introducing the counterfactual reasoning, as well as its broad applicability to various existing  KT models.

\begin{table}[t]
\centering
\caption{Performance comparison between different methods on \emph{the unbiased test sets} of three benchmarks datasets.
The bold-face font denotes the winner in that column.
% Note that the improvement achieved by our CORE framework is statistically significant at \bm{$p < 0.05$}.
}
\begin{tabular}{l c c c c c c c c}
\toprule
\multirow{2}{*}{} & \multicolumn{2}{c}{ASSIST09} & & \multicolumn{2}{c}{ASSIST17} & & \multicolumn{2}{c}{EdNet} \\
\cmidrule{2-3}\cmidrule{5-6}\cmidrule{8-9}
& Accuracy & AUC & & Accuracy & AUC & & Accuracy & AUC \\
\midrule
% \cmidrule{2-3}\cmidrule{5-6}\cmidrule{8-9}
BKT\rule{0mm}{3mm} & 50.09\% & 50.26\% & & 52.38\% & 54.26\% & & 51.30\% & 52.90\% \\
SAKT\rule{0mm}{3mm} & 59.93\% & 64.86\% & & 56.30\% & 58.06\% & & 52.75\% & 55.17\% \\
SAINT\rule{0mm}{3mm} & 62.17\% & 68.71\% & & 55.39\% & 58.87\% & & 52.37\% & 54.81\% \\
\midrule
DKT\rule{0mm}{3mm} & 58.40\% & 62.88\% & & 60.98\% & 66.43\% & & 51.43\% & 52.80\% \\
DKT-CORE\rule{0mm}{3mm} & 64.24\% & 69.23\% & & \textbf{65.16\%} & \textbf{71.06\%} & & \textbf{55.67\%} & \textbf{57.57\%} \\
\midrule
DKVMN\rule{0mm}{3mm} & 61.11\% & 66.66\% & & 61.19\% & 67.13\% & & 52.60\% & 56.13\% \\
DKVMN-CORE\rule{0mm}{3mm} & 63.36\% & 68.76\% & & 62.95\% & 68.30\% & & 54.88\% & 57.37\% \\
\midrule
AKT\rule{0mm}{3mm} & 62.05\% & 68.63\% & & 59.64\% & 65.94\% & & 52.48\% & 56.74\% \\
AKT-CORE\rule{0mm}{3mm} & \textbf{64.51\%} & \textbf{70.02\%} & & 61.81\% & 66.71\% & & 54.56\% & 57.12\% \\
\bottomrule
\end{tabular}
\label{tbl:unbiased_performance}
\end{table}

As can be clearly seen, the traditional BKT performs the worst among all baselines, and DKT lags behind the other deep learning based baselines by a large margin.
% The reason behind this may be two-folds: 1) KT models with simple architectures have difficulty capturing the complex dependencies among students' learning interactions; and 2) Simpler KT are more prone to the spurious correlations between questions and students' responses, i.e., the answer bias, making them more vulnerable to the elimination of the bias.
A possible reason is that simple KT models may have difficulty capturing the complex dependencies among students' learning interactions
On the other hand, our CORE framework appears to provide larger gains for simple KT models.
For instance, the advantage of DKT-CORE over DKT is more noticeable than that of DKVMN-CORE over DKVMN and AKT-CORE over AKT.
In four out of the six metrics, DKT-CORE even takes the first place.
This means that when the backbone model has a relatively weak learning ability, our CORE framework can effectively compensate for this limitation to make the debiased inference.

\begin{figure}[t]
    \centering
    \subfloat[DKT-CORE]
    {\includegraphics[width=0.325\textwidth]{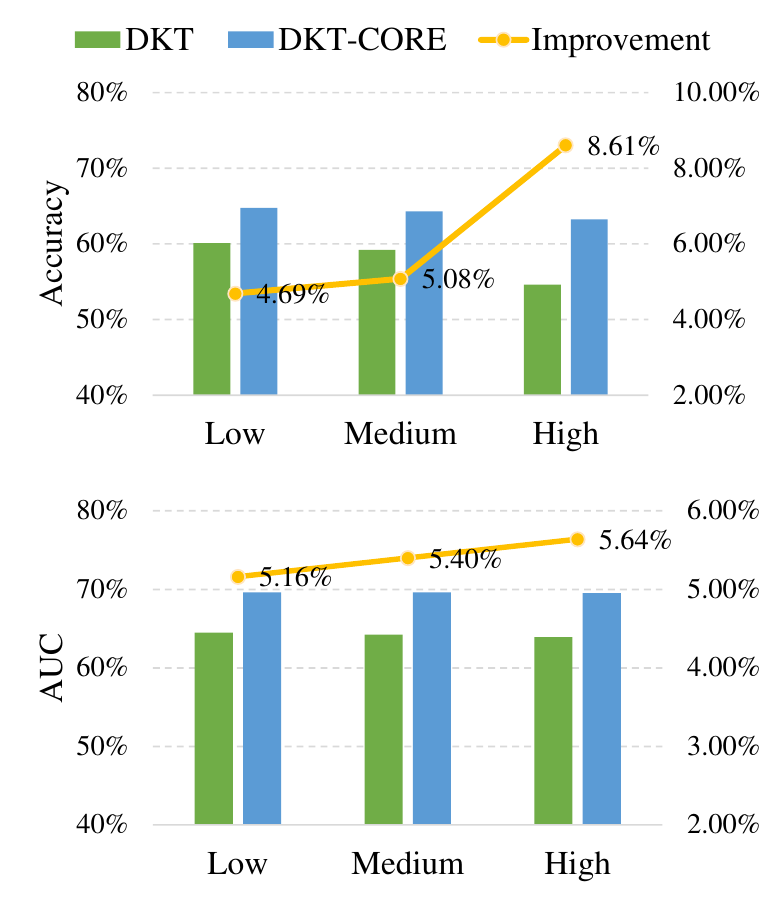}}
    \hspace{0.005\textwidth}
    \subfloat[DKVMN-CORE]
    {\includegraphics[width=0.325\textwidth]{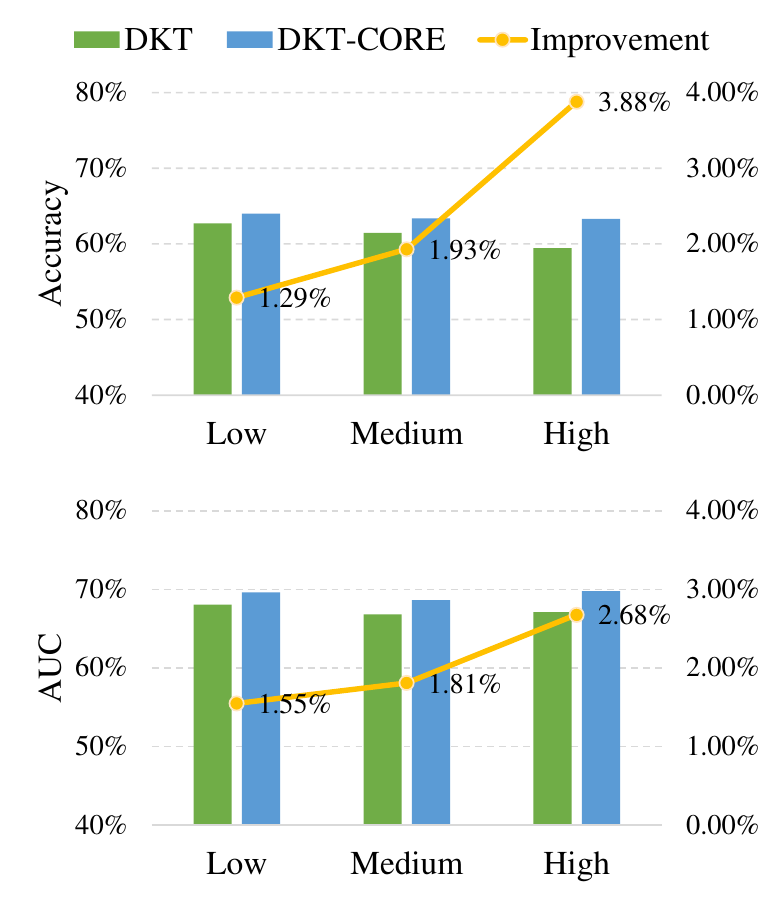}}
    \hspace{0.005\textwidth}
    \subfloat[AKT-CORE]
    {\includegraphics[width=0.325\textwidth]{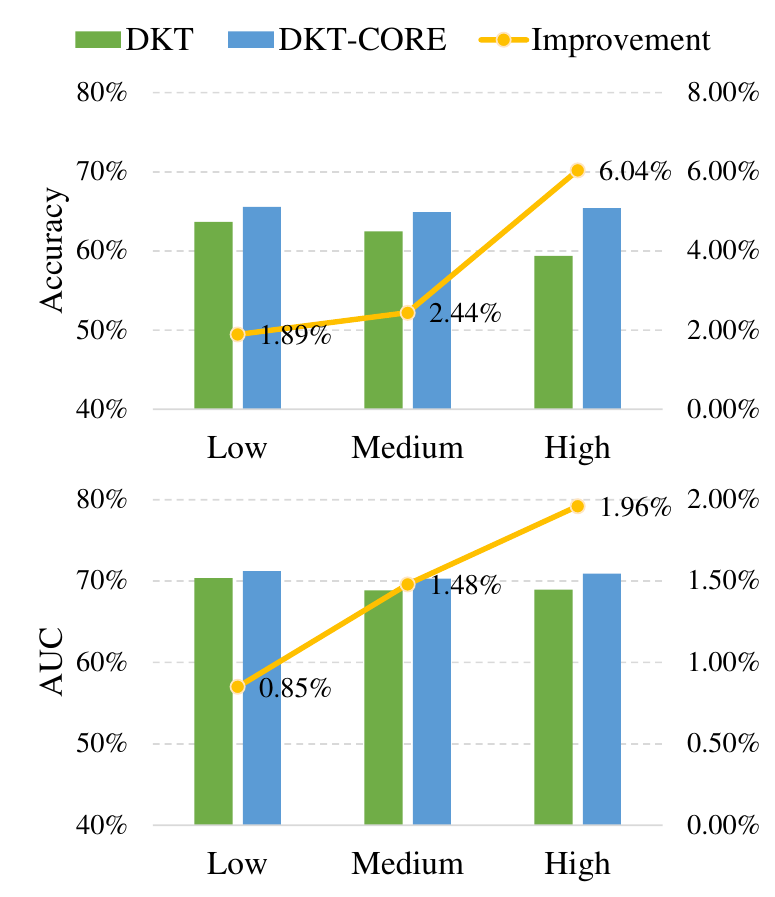}}
     \caption{Performance advantage of our CORE framework for questions with different bias strengths on ASSIST09. The method performance is plotted against the left axis, while the corresponding improvement is indicated along the right axis.}
    \label{fig:group_performance}
\end{figure}

In addition, we examine how CORE performs for questions with different strengths of answer bias.
We measure the bias strength for a question by the ratio of the more frequent answer (either correct or incorrect) to the total number of answers in the training set\footnote{The strength of answer bias ranges from $0.5$ to $1.0$.}.
We divide all questions into three groups according to their bias strengths: low (questions with a bias strength lower than 0.6), medium (questions with a bias strength between 0.6 and 0.8), and high (questions with a bias strength higher than 0.8).
Fig.~\ref{fig:group_performance} shows the performance advantage of CORE across various question groups on ASSIST09.
As expected, DKT-CORE, DKVMN-CORE, and AKT-CORE consistently outperform their respective backbone models for each group.
More importantly, we observe that the advantage of CORE generally becomes larger as the bias strength of questions increases.
DKT-CORE, DKVMN-CORE, and AKT-CORE show the most significant improvement for the questions with high bias strength, achieving an increase of 8.61\%, 3.88\%, and 6.04\% in accuracy, respectively.
The phenomenon demonstrates that the effectiveness of CORE indeed stems from its strong ability to reduce the impact of answer bias on KT.

\begin{table}[t]
\centering
\caption{Ablation study by comparing the performance of DKT-CORE and its variants.}
\begin{tabular}{l c c c c c c c c}
\toprule
\multirow{2}{*}{} & \multicolumn{2}{c}{ASSIST09} & & \multicolumn{2}{c}{ASSIST17} & & \multicolumn{2}{c}{EdNet} \\
\cmidrule{2-3}\cmidrule{5-6}\cmidrule{8-9}
& Accuracy & AUC & & Accuracy & AUC & & Accuracy & AUC \\
\midrule
% \cmidrule{2-3}\cmidrule{5-6}\cmidrule{8-9}
DKT\rule{0mm}{3mm} & 58.40\% & 62.88\% & & 60.98\% & 66.43\% & & 51.43\% & 52.80\% \\
Variant with TE\rule{0mm}{3mm} & 62.16\% & 67.51\% & & 61.68\% & 68.71\% & & 52.26\% & 55.42\% \\
Variant with $p=0$\rule{0mm}{3mm}  & 63.79\% & 68.90\% & & 64.30\% & 70.41\% & & 55.23\% & 57.30\% \\
Variant without ${\mathcal{L}_{BCE}^{q}}$\rule{0mm}{3mm} & 62.67\% & 68.19\% & & 61.64\% & 68.87\% & & 53.57\% & 56.05\% \\
DKT-CORE\rule{0mm}{3mm} & \textbf{64.24\%} & \textbf{69.23\%} & & \textbf{65.16\%} & \textbf{71.06\%} & & \textbf{55.67\%} & \textbf{57.57\%} \\
\bottomrule
\end{tabular}
\label{tbl:variant_performance}
\end{table}

\subsection{Ablation Study}
To investigate the contributions of some key designs in our CORE framework, we carried out ablation studies by comparing the performance of DKT-CORE and its variants.
The results are shown in Table~\ref{tbl:variant_performance}.

In CORE, we argue that the impact of answer bias lies in the direct effect of questions on students' responses.
As a result, we make the debiased inference for KT by subtracting the direct effect from the total effect, i.e., $TE-NDE$ as shown in Eq.~\eqref{eq:TIE}.
DKT-CORE is first compared against its variant that directly uses TE without subtracting NDE for inference.
As can be seen, the variant method using TE is substantially inferior to DKT-CORE across all datasets.
This observation verifies our hypothesis that the impact of answer bias is indeed hidden in the direct effect of questions and can be mitigated by reducing it.
Moreover, the variant method is still ahead of the original baseline model DKT in most cases.
One possible reason is that when calculating TE, the students' individual factors that influence their responses can be considered through the causal path $S \to R$, which, however, is entirely neglected in DKT.

Besides, NDE needs to be estimated in the counterfactual KT, where both the values of student and knowledge state are set to null values.
In this situation, we introduce the model parameter $p$ as the outputs of the student-only branch and student-question branch.
We optimize $p$ in CORE by imposing a KL loss that requires the prediction in the counterfactual KT to be consistent with that in the conventional KT as shown in~Eq.~\eqref{eq: loss_kl}.
This step can be skipped by adopting a predetermined $p$.
Therefore, we implement a variant of DKT-CORE by simply setting $p=0$, assuming that the network branches produce zero outputs if the inputs are null values.
From Table~\ref{tbl:variant_performance}, we can see that the variant method with predetermined $p$ falls behind DKT-CORE, implying that it is preferable to determine $p$ in a learning manner.
Meanwhile, the variant method is far better than DKT even though the estimation of NDE is simplified, which proves the robustness of CORE to some extent.

To better model the direct impact of questions on students' responses, we incorporate an additional BCE loss ${\mathcal{L}_{BCE}^{q}}$ as shown in Eq.~\eqref{eq: loss_q}.
We compare DKT-CORE with a variant that does not include ${\mathcal{L}_{BCE}^{q}}$, and observe a clear drop in performance of the latter relative to the former.
This phenomenon demonstrates the potential benefits of incorporating ${\mathcal{L}_{BCE}^{q}}$ in our CORE framework.

\subsection{Case Study}
\begin{figure}
    \centering
    \includegraphics[width=0.5\textwidth]{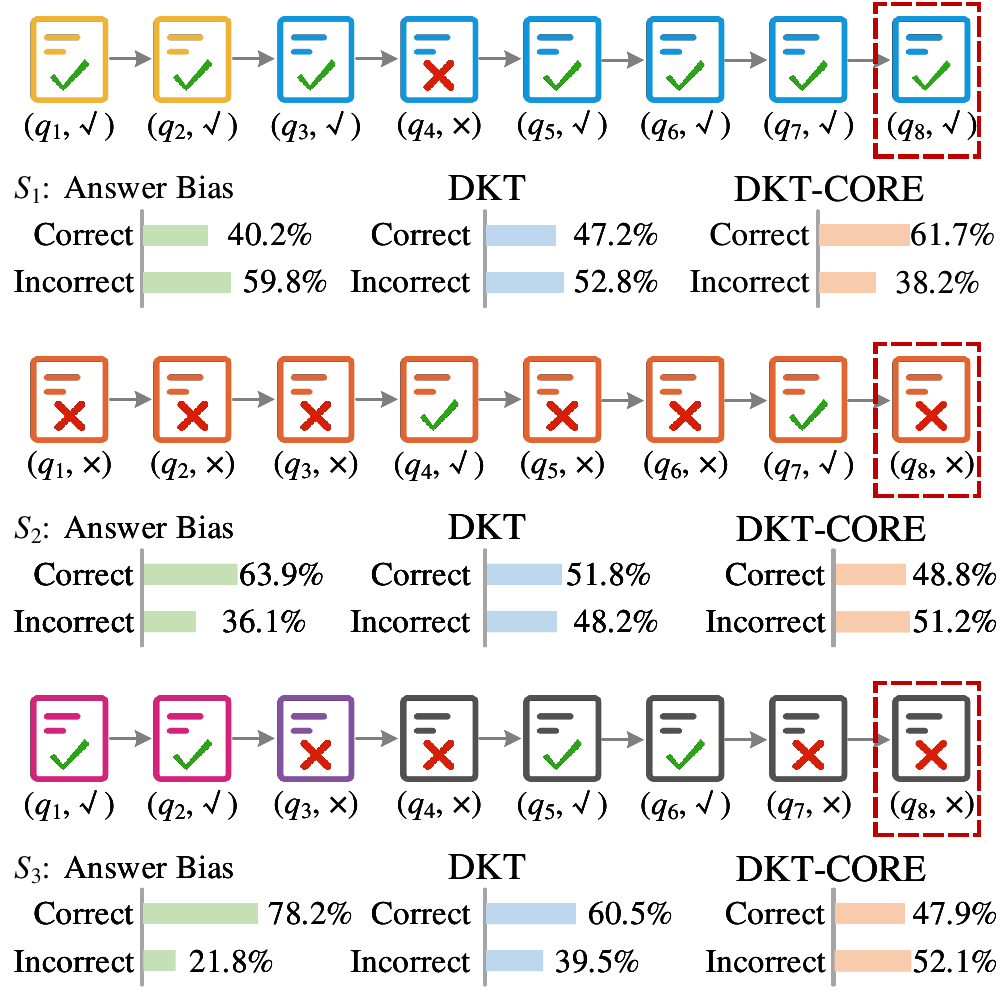}
    \caption{A case study of the learning sequences of three test students $s_1$, $s_2$, and $s_3$, and the probabilities predicted by DKT and DKT-CORE that they will correctly and incorrectly answer the last question $q_8$ in the red dotted box. The questions related to the same concept are marked with the same color.}
    \label{fig:case_study}
\end{figure}

We perform a case study by visualizing the learning sequences of three test students, denoted as $s_1$, $s_2$, and $s_3$ in Fig.~\ref{fig:case_study}.
The questions related to the same concept are marked with the same color.
We show the probabilities predicted by DKT and DKT-CORE that the students will correctly and incorrectly answer the last question $q_8$ in the sequence, as well as the bias between correct and incorrect answers for the question.
For the student $s_1$, the answer bias suggests that students tend to make mistakes more often for the question $q_8$.
DKT agrees with this logic and predicts that $s_1$ is more likely to answer incorrectly.
However, in recent learning interactions, $s_1$ has consistently provided correct answers to the questions that are associated with the same concept as $q_8$.
Therefore, it is more reasonable to expect that $s_1$ will answer $q_8$ correctly, which indeed is the case.
In contrast, DKT-CORE gives the accurate prediction, showcasing its resistance to answer bias.
Similarly, since DKT-CORE can concentrate more on the understanding of students' knowledge states, it avoids the misprediction of DKT for the student $s_2$, which is also caused by the impact of answer bias.

For the student $s_3$, the answer bias implies that the question $q_8$ is relatively straightforward, and hence DKT blindly predicts that the probability of correctly answering is much higher than that of incorrectly answering.
On the other hand, DKT-CORE effectively mitigates the influence of answer bias and recognizes that despite of the simplicity of $q_8$, $s_3$ has repeatedly made mistakes in the past.
This reveals that $s_3$ may not have a firm grasp of the corresponding knowledge concept.
As a result, DKT-CORE makes the right prediction that $s_3$ will answer $q_8$ incorrectly.
These examples provide us with intuitive insights into the advantages of DKT-CORE.

\section{Conclusions}\label{sec:conclusions}
KT is a critical means to support various intelligent educational services.
In this paper, we have demonstrated that the success of many recently proposed KT models is largely driven by memorizing the answer bias present in datasets.
They do not fully understand students' knowledge states, which is the real goal of KT.
This issue has not been adequately discussed in the literature, and we hope that our work will inspire new research endeavors to tackle this important problem.

As an initial attempt, we have approached the KT task from a causality perspective and identified the impact of answer bias as the direct causal effect of questions on students' responses.
A counterfactual reasoning framework CORE has been presented to estimate different causal effects and mitigate answer bias by subtracting the direct effect from the total effect.
Notably, CORE is a model-agnostic framework, and we have implemented it based upon various prevalent KT models.
We have established an unbiased evaluation to reflect the true ability of models in knowledge state understanding.
Experimental results on benchmark datasets have verified the effectiveness of CORE.

In the future, we intend to explore more sophisticated modules, instead of the basic multi-layer perceptron, to better instantiate the student-only branch and question-only branch.
Moreover, we would like to incorporate additional side information about questions and students for KT.
This includes considering the knowledge structure of questions, as well as the response time and opportunity counts of students during the learning process.
Finally, most existing KT methods only focus on estimating binary responses (i.e., correctness or incorrectness) of students.
However, in reality, students' learning interactions are much more diverse and complicated.
Therefore, we plan to extend our method beyond correctness prediction.
For example, we shall predict students' exact options on multiple-choice questions.



\bibliographystyle{elsarticle-num}
\bibliography{COREKT_Ref}


\end{document}